\documentclass{article}
\usepackage{spconf,amsmath,graphicx}
\usepackage[most]{tcolorbox}

\title{Residual Pyramid FCN for Robust Follicle Segmentation}
\name{\normalsize{Zhewei Wang$^1$ \qquad Weizhen Cai$^1$\qquad Charles
    D. Smith$^2$ \qquad Noriko Kantake$^3$ \qquad Thomas J. Rosol$^3$ \qquad
    Jundong Liu$^1$\footnote{Corresponding
      author. Email}}
}
\address{\normalsize{$^1$School of EECS, Ohio University, Athens, OH 45701} \\
  $^2$ \normalsize{Department of Neurology, University of Kentucky,
    Lexington, KY 40508} \\
  $^3$ \normalsize{Department of Biomedical Sciences,
    Ohio University,
    Athens, OH 45701} }

\begin{document}
%
\maketitle
\begin{abstract}

  In this paper, we propose a pyramid network structure to improve the
  FCN-based segmentation solutions and apply it to label thyroid
  follicles in histology images. Our design is based on the notion
  that a hierarchical updating scheme, if properly implemented, can
  help FCNs capture the major objects, as well as structure details in
  an image. To this end, we devise a residual module to be mounted on
  consecutive network layers, through which pixel labels would be
  propagated from the coarsest layer towards the finest layer in a
  bottom-up fashion. We add five residual units along the decoding
  path of a modified U-Net to make our segmentation network,
  Res-Seg-Net. Experiments demonstrate that the multi-resolution
  set-up in our model is effective in producing segmentations with
  improved accuracy and robustness.

\end{abstract}

\section{Introduction}

Thyroid is the largest endocrine gland in human body and it produces
hormones that influence the metabolic rate and protein
synthesis. Follicles make up one of the major components of thyroid
glands. The morphology of follicle cells can often serve as a reliable
indicator of the health status of the glands –- healthy cells are
usually homogeneous while cell polymorphism likely signals an abnormal
mutation, from inflammations to cancers \cite{wang2010detection}.
Therefore, Identification and evaluation of follicle polymorphism
through histological images are of great importance for thyroid cancer
diagnosis, as well as treatment planning.

Separating follicles from the surrounding tissue is often a
prerequisite step for many other analysis tasks. As manual
delineations are normally tedious, time-consuming and prone to intra-
and inter-operator errors, various automatic solutions have been
proposed in the past 20 years or so. Traditional approaches include
boundary tracing \cite{borst1979thresholding}, 
watershed \cite{adiga2006high}, graph cuts
\cite{boykov2001interactive}, 
and Gaussian mixture models \cite{liu_2009_mmbia}. They commonly take
certain hand-crafted features, e.g., edges or texture, as the basis
for pixel labeling and subsequent analysis.

In recent years, deep neural networks have emerged as a new and more
powerful paradigm, which revolutionized many artificial intelligence
areas, including semantic segmentation. Fully convolutional network
(FCN) \cite{long2015fully} and its variants, including U-Net
\cite{ronneberger2015u}, produce state-of-art results on many data and
applications. The success of FCNs should be attributed, in great part,
to their capability of processing input images at different spatial
scales. FCNs are commonly constructed with an encoder-decoder
architecture. In the encoding path, input images are processed through
a number of convolution + pooling layers to generate high-level latent
features, which are then progressively upsampled in the decoder to
reconstruct the target pixel labels.

A crucial issue in FCN design is how to effectively integrate the
feature maps produced in higher (finer) layers and those in lower
(coarser) layers. The former are richer in semantics, while the latter
carry more spatial details that define class boundaries. Early efforts
lead to the developments of skip architecture \cite{long2015fully},
bridges with feature concatenation \cite{ronneberger2015u}, dilated
convolution \cite{yu2017dilated}, up-down-sampling
\cite{zhang2018exfuse}, among others. New models have emerged in the
past two years, such as the utilizations of dilated pooling
\cite{sarker2018slsdeep}, supervision with additional labels
\cite{zhang2018deep}, multi-task learning \cite{playout2018multitask},
multi-view ensemble \cite{chen_isbi_2017}, convolutional LSTM
\cite{chen_mlmi_2017} and shape preserving loss
\cite{yan2018deep}. Originated from the first FCN, most existing
solutions follow a common setup: the network objective function is
defined at the last layer, between its outputs and the ground-truth
masks. While the features are learned and propagated along layers in a
multi-scale manner, their updates are solely driven with an overall
penalty defined on the final outputs. Hierarchical updates through
image pyramids, proven effective in many previous studies
\cite{burt1987laplacian}, are essentially lacking in most of the
existing FCN models. Zhang {\it et al.}  \cite{zhang2018deep} utilizes
a multi-resolution loss, but the computation is still not conducted
within a sequential update framework.

To seek a remedy, we propose a new FCN-based model in this paper and apply it to thyroid follicle segmentation. The design goal is to equip FCNs with a level-by-level hierarchical updating mechanism, hoping it will lead to more robust and accurate segmentation performance. To this end, we include additional loss terms based on low-layer feature maps to ensure a good starting segmentation at coarse levels. We also employ residual units to facilitate level-by-level segmentation refinements, when network inference is carrying out along the decoding path. With these two setups, our model can take full advantage of hierarchical multi-resolution processing.

\section{Method}
Many traditional image analysis solutions \cite{marfil2006pyramid,
  thevenaz1998pyramid} have demonstrated that multi-resolution
representations \cite{burt1987laplacian} enable effective processing
pipelines for both segmentation and registration tasks. In these
models, input images are resampled and transformed into coarse levels
in a bottom-up manner. The processed results are then propagated in a
reverse top-down direction to provide the fine levels with better
starting estimations, often leading to more accurate final results.

However, such bottom-up and top-down input/result transitions are
absent in FCNs. No intermediate segmentation result at any coarse
resolution is generated in FCNs to serve as a guidance for finer
levels. The loss functions in most FCNs are defined only at their
final layers, between the network predictions and ground-truth
segmentations. With such losses, which put no emphasis on coarse level
results, the efficacy of an end-to-end learning, especially through a
network with many layers, may be greatly hindered.

These observations lead us to the development of a new FCN model. We
modify the decoding layers of U-Net with an intention to impose a
bottom-up structure, through which the segmentations from coarse
levels can be transited and refined in fine layers. To further ensure
the refinements to take place in an effective way, we adopt the
residual concept \cite{he2016deep} to design a residual module as the
building block for our network. We term this module Res-Seg.

\subsection{Res-Seg module}

\begin{figure}[htb]

\begin{minipage}[b]{1.0\linewidth}
  \centering
  \centerline{\includegraphics[width = 8.5cm]{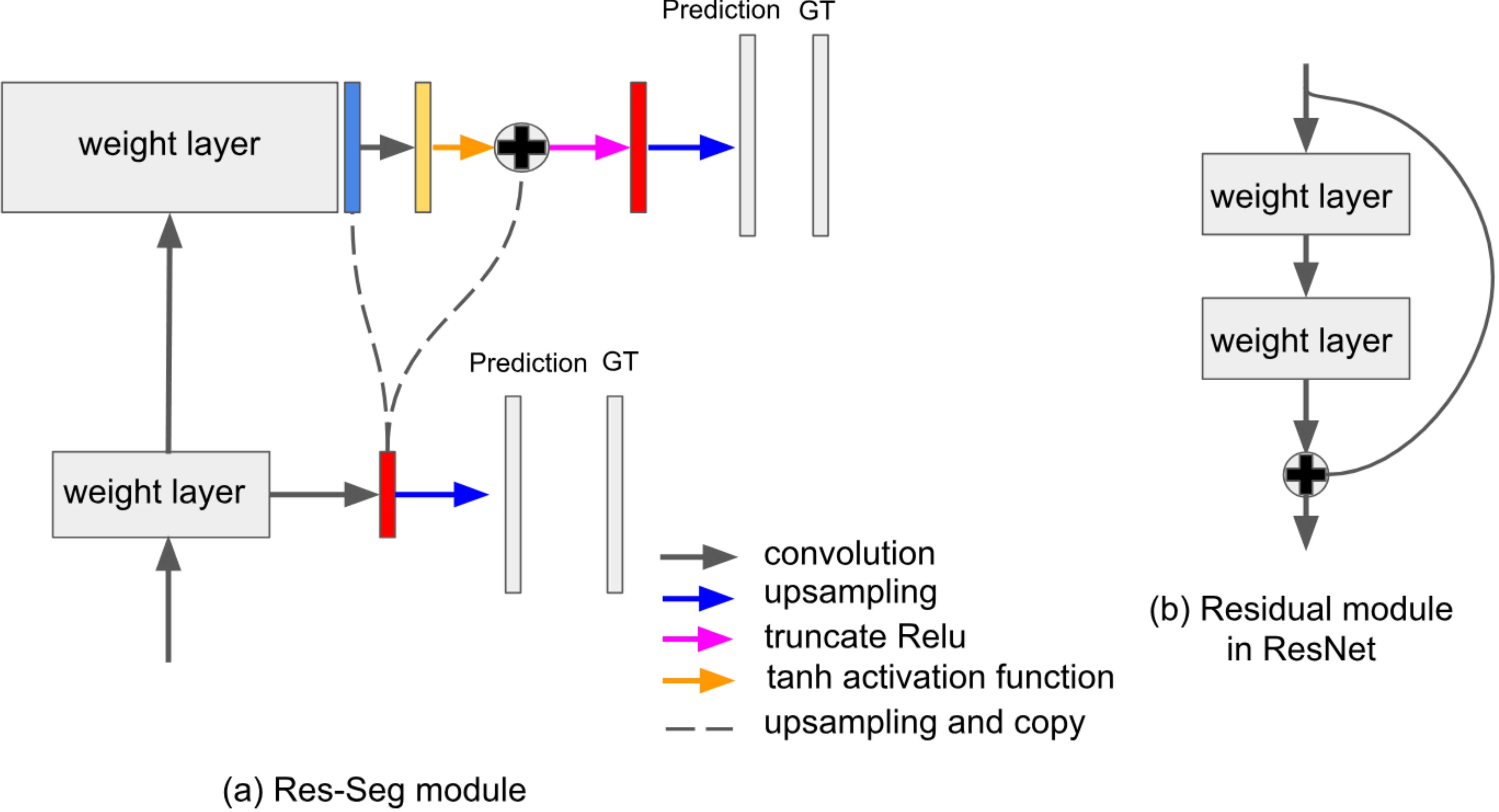}}
\end{minipage}

\caption{(a) Res-Seg module for segmentation. Picture is best viewed in color. (b) Residual blocks proposed in ResNet \cite{he2016deep}.}
\label{fig:res-seg}
\end{figure}

Fig.~\ref{fig:res-seg} (a) shows a Res-Seg unit of two consecutive
layers. At each layer, an intermediate probability map (prob-map),
shown as a red bar, is generated and compared (after upsampling) with
the ground-truth segmentation. Meanwhile, the prob-map from a low
resolution layer is sent to its upper (finer) layer with two aims: 1)
it will provide a good starting point for the latter; 2) it will get
refined in the upper layer. To facilitate the finer layer to focus on
learning the increment details, we adopt a similar residual-connection
structure from the ResNet \cite{he2016deep} for our Res-Seg module.

As shown in Fig.~\ref{fig:res-seg} (a), the prob-map from the low
level is upsampled (blue bar) and then sent to the high layer at two
spots. Firstly, the blue bar is concatenated with the feature maps to
help produce the prob-map at the finer level. The second location is
on the path from the concatenated feature/prob maps to their own
prob-map at the upper layer. With that, the fine layer essentially
learns the residual (yellow bar) between the upsampled coarse and
ground-truth segmentations. In other words, the fine layer is
positioned to learn refinements for the coarse segmentation.

We have two considerations in the design of this residual unit. Firstly, the upsampled lower layer prob-map, the blue bar, has the values in the range of [0, 1]. The direct output of the upper layer, the yellow bar, which is the residual between the blue bar and the ground-truth, is in the range of [-1, 1]. With this consideration, tanh is a good choice for the activation function to be applied on the concatenated feature/prob maps. This operation is shown as a yellow arrow in Fig.~\ref{fig:res-seg}. The second consideration is about the activation function in generating the prob-map at the upper layer. In this work, we chose a truncated ReLU, $f(x) = \max(\min(x, 1), 0)$, to map the output of this layer into the range of [0, 1].

While our Res-Seg module is inspired by the residual layers in ResNet, there are several major differences in their designs and functionalities. Firstly, our Res-Seg module is designed for segmentation and the output of each layer is supervised by the ground truth. The ResNet blocks in Fig.~\ref{fig:res-seg} (b), however, are not directly regulated by the ground-truth class labels. The purpose of ResNet blocks is to propagate the context information to the next layer, while our Res-Seg module aims to bring a direct guidance from the ground-truth at each layer. The second major distinction lies in the structures of the two modules. As illustrated in the last paragraph, there are several well-grounded connections and activation function setups in our res-seg, designed specifically for semantic segmentation.

{\bf Two updating schemes} To ensure each upper layer to only focus on refining the results (prob-maps) sent from the previous layer, these prob-maps should be fixed at both plugging spots in the upper layers. This setup would conform to the design goal of our level-by-level updating scheme. In our implementation, the prob-map of each layer along the decoding path is copied to its upper layer, and set to untrainable there. By doing so, we force the higher layer learn the differences between the lower layer prob-map and the ground-truth. A different setting, however, can be borrowed from the ResNet, where all layers/weights are updated through backpropagation. While deviating from our level-by-level design notion, this approach certainly grants more flexibility over the learning procedure, allowing the network to minimize the total loss to its full potential. In this work, we implement both settings to compare their performance.

It should be noted that our residual module is rather general, and it can be integrated into the decoding path of many FCN networks. We choose U-Net as the baseline network for comparison and modification. We term our overall segmentation network as Res-Seg-Net.

\subsection{Res-Seg-Net}
We keep the overall architecture of the original U-Net, including the number of layers, in our Res-Seg-Net. Two major modifications have been made on U-Net to fit our data and task. Firstly, we reduce the number of channels at each layer, to have fewer parameters. We start with 32 channels (64 channels in U-Net), followed by doubling the number of channels at each down-sampling step along the contraction path. The expansion path is kept symmetric to the contraction path. Secondly, we use padding in every convolution operation to maintain the spatial dimension. Dimension changes only occur at polling (reduced to half) and upsampling (enlarge two times). In this way, we do not have to crop the contraction layers, as U-Net does, for skip connections.

\begin{figure}[htb]

\begin{minipage}[b]{1.0\linewidth}
  \centering
  \centerline{\includegraphics[width = 8.5cm]{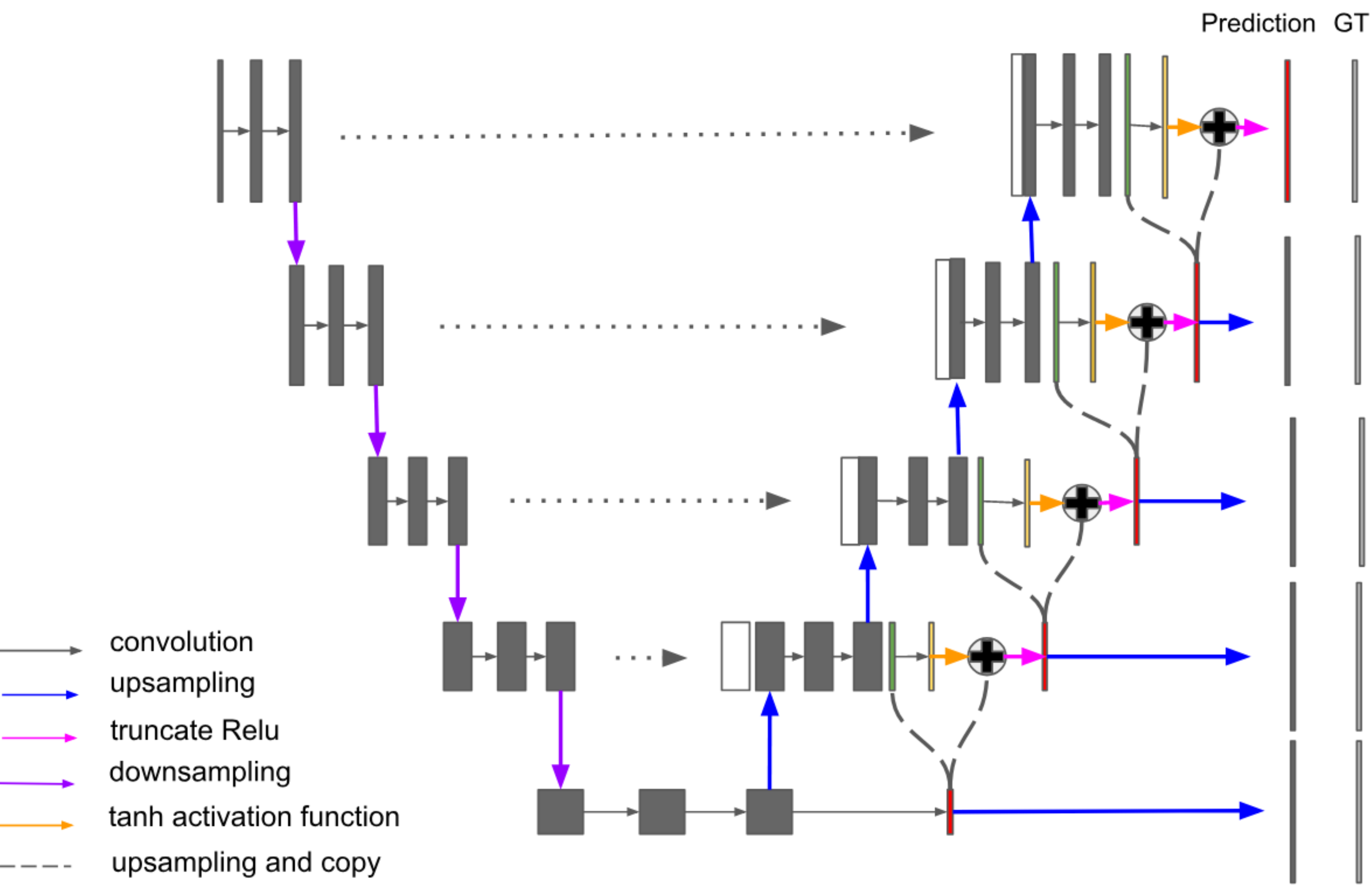}}
\end{minipage}

\caption{Res-Seg-Net architecture. $GT$ stands for ground-truth.}
\label{fig:res-u-net}
\end{figure}

The integration of modified U-Net with our proposed Res-Seg module starts at the bottom (lowest resolution) layer of the network. Following the expansion path, our residual module is applied on each pair of adjacent upsampling layers, as shown in Fig.~\ref{fig:res-u-net}. Overall, the network produces four intermediate and one final probabilistic segmentation maps. At each layer, the intermediate segmentation map is obtained by upsampling the respective prob-map, through a single bilinear filter, to match the dimension and resolution of the ground-truth mask.

{\bf Multi-resolution loss function} We resort to Dice loss \cite{vnet_2016} to measure the difference between each segmentation map with the ground-truth mask. The choice is based on the facts: 1) Dice Similarity Coefficient (DSC) is a common metric to evaluate segmentation performance, which is also adopted in this work; 2) Dice loss has been widely used as a differentiable approximation of DSC.
Let $S$ be the segmentation result produced by a solution and $R$ be the ground truth.  In Dice loss, segmentation $S$ is relaxed to a probability map of real numbers between 0 and 1, and the loss is computed as:
\begin{equation} \label{eq:1}
\textrm{Dice loss} = -\frac{2\sum_i{s_ir_i}}{\sum_is_i+\sum_ir_i}
\end{equation}
where $s_i\in[0,1]$ is the label prediction at pixel $i$, and
$r_i\in \{0,1\}$ is the corresponding binary ground truth. The overall object function in our Res-Seg-Net is defined as the
summation of weighted Dice losses, which is
\begin{equation} \label{eq:2}
{L_{Dice}} = -\sum_iw_iD_i
\end{equation}
where $D_i$ is the Dice loss of an individual layer, and $w_i$ is the
corresponding weight. In our experiment, the weight of the Dice loss
is set as 1/4 for intermediate layers, and 1 for the final layer. We
give the highest resolution layer larger weight as it produces the
ultimate segmentation prediction of the network.


\section{Experiments}

{\bf{Data}} To evaluate the effectiveness of our proposed Res-Seg-Net for follicle segmentation, we conducted experiments on histology slides of rat thyroid stained with hematoxylin-eosin (H \& E). The images have resolution of 1$\mu$m/pixel. Ground-truth segmentations were generated based on manual delineations. A veterinary pathologist traced the boundaries of all distinguishable follicles and colloids in one particular lobe. QuPath \cite{bankhead2017qupath} was employed to trace continuous contours, which are actually polygons. As our network processes and outputs image matrices, we converted the ground-truth segmentation from polygons to binary masks through the point-in-polygon algorithm \cite{shimrat1962algorithm}.

Fixed-sized square subimages (800 $\times$ 800) were randomly sampled from
the slides. Totally 100 such subimages were extracted from the
original pathology slides, and together with the manual masks, they
make up our data in this work. To validate our models, the 100
image-mask pairs were randomly separated into training, validation and
test groups, with a size ratio of 8:1:1. In order to obtain
more training samples, as well as to reduce overfitting, we further
extracted smaller-sized (640 $\times$ 640) patches from the subimages, also
in a random manner.  The patches are also arbitrarily flipped
to augment the training data.  The validation set
is used to select optimal hyper-parameters in our models (which is epoch number of early stopping).

\begin{figure}[htb]
\centering
\begin{tcbraster}[raster columns=3, raster equal height,
raster column skip=10pt, raster row skip=4pt, raster every box/.style={blank}, width=5cm]
\tcbincludegraphics{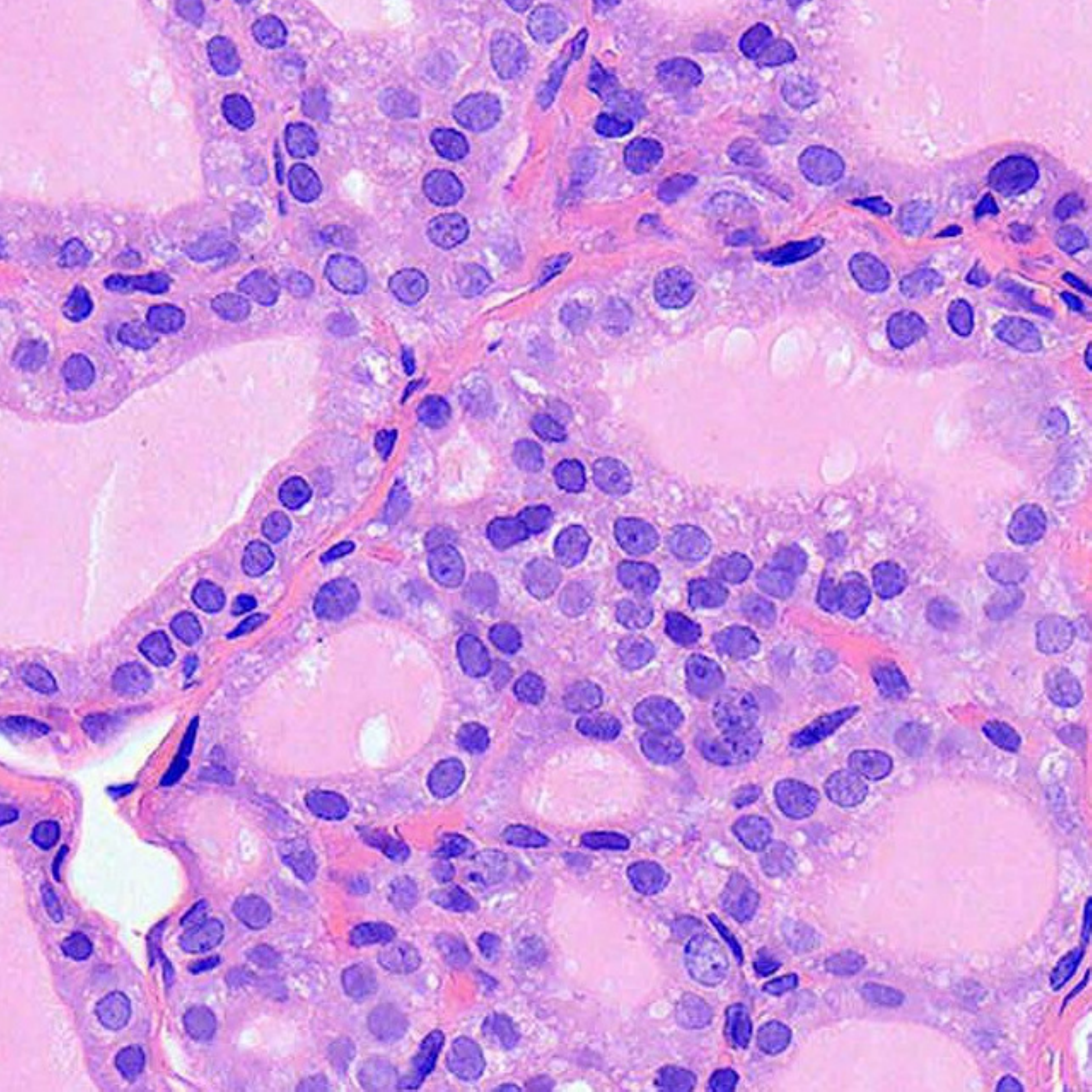}
\tcbincludegraphics{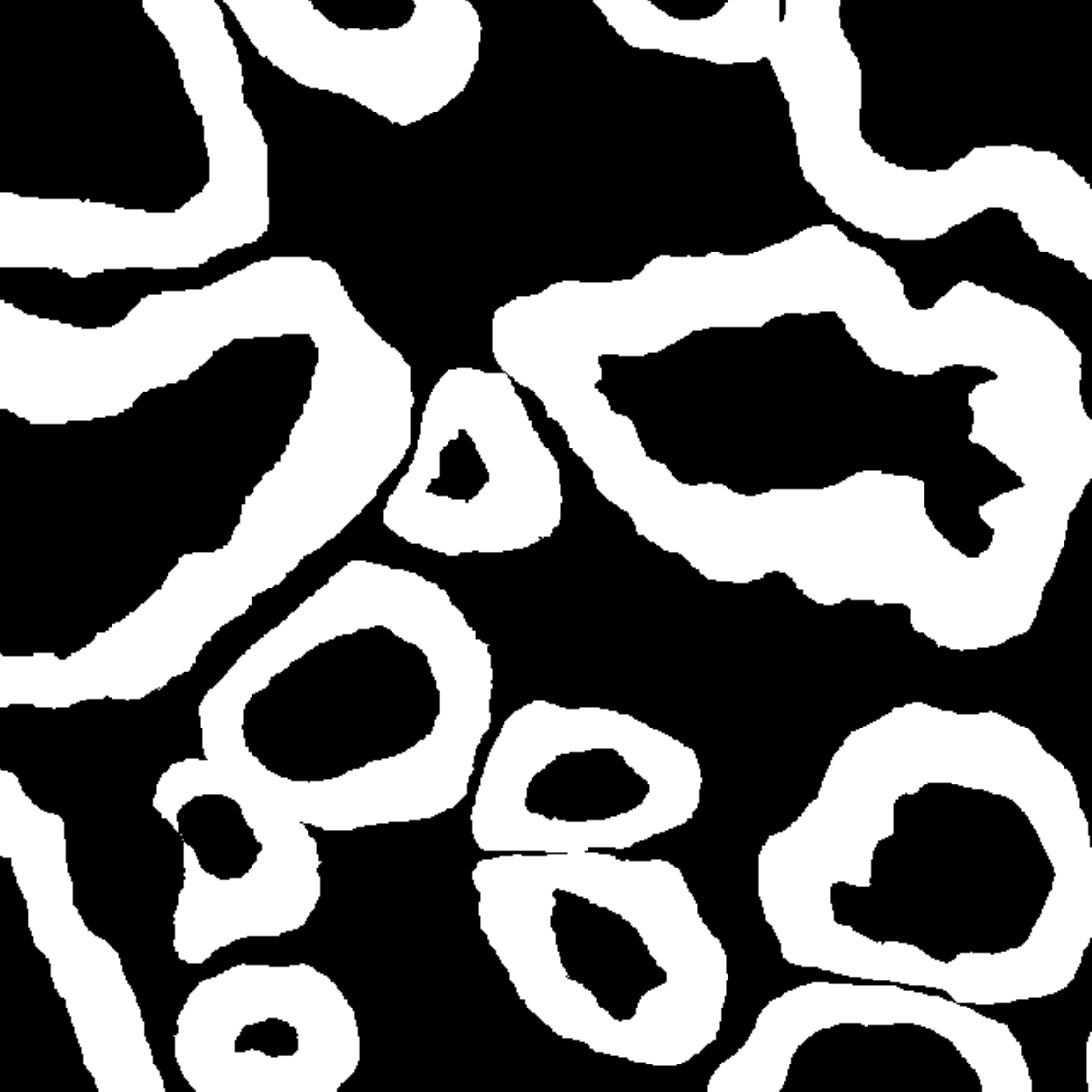}
\tcbincludegraphics{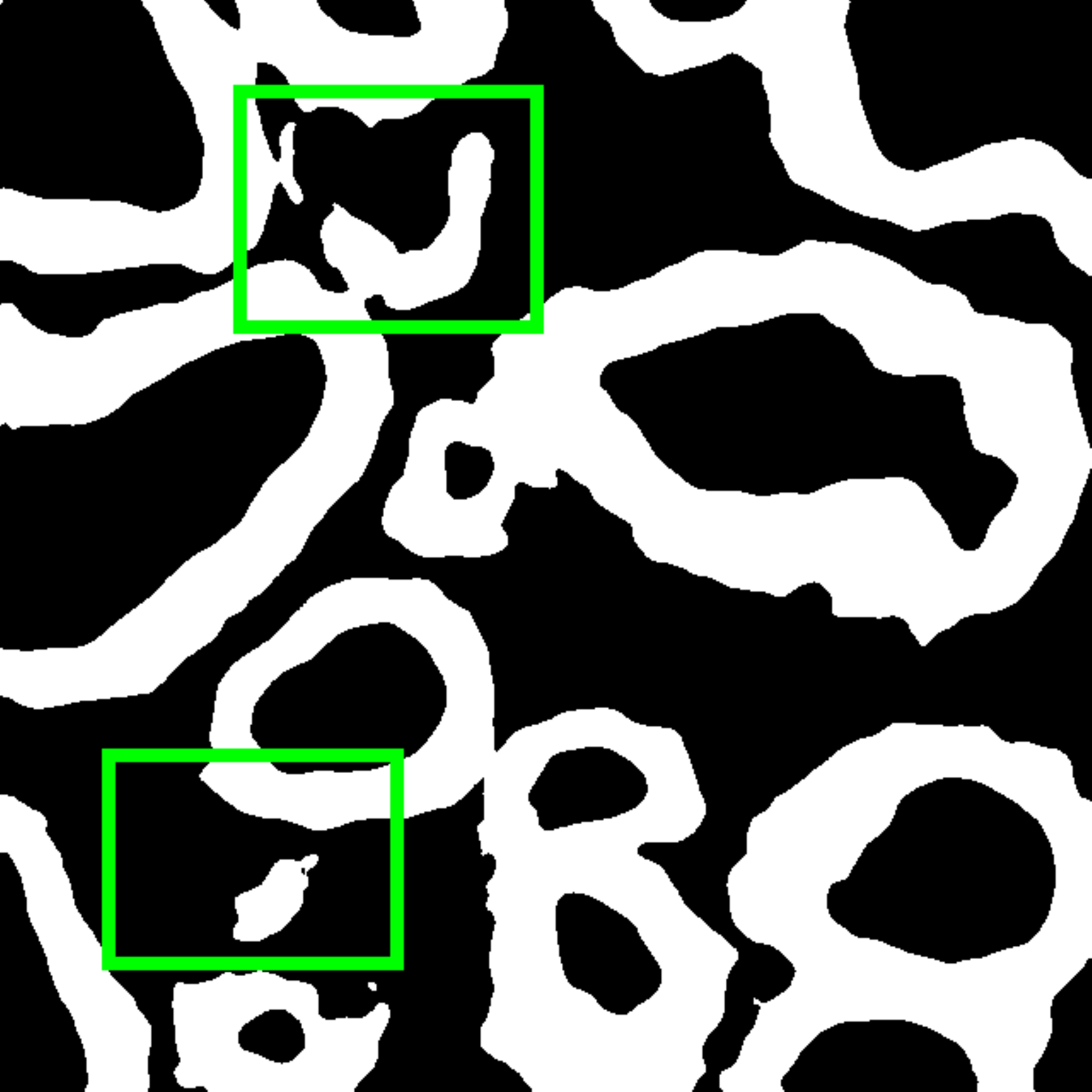}
\tcbincludegraphics{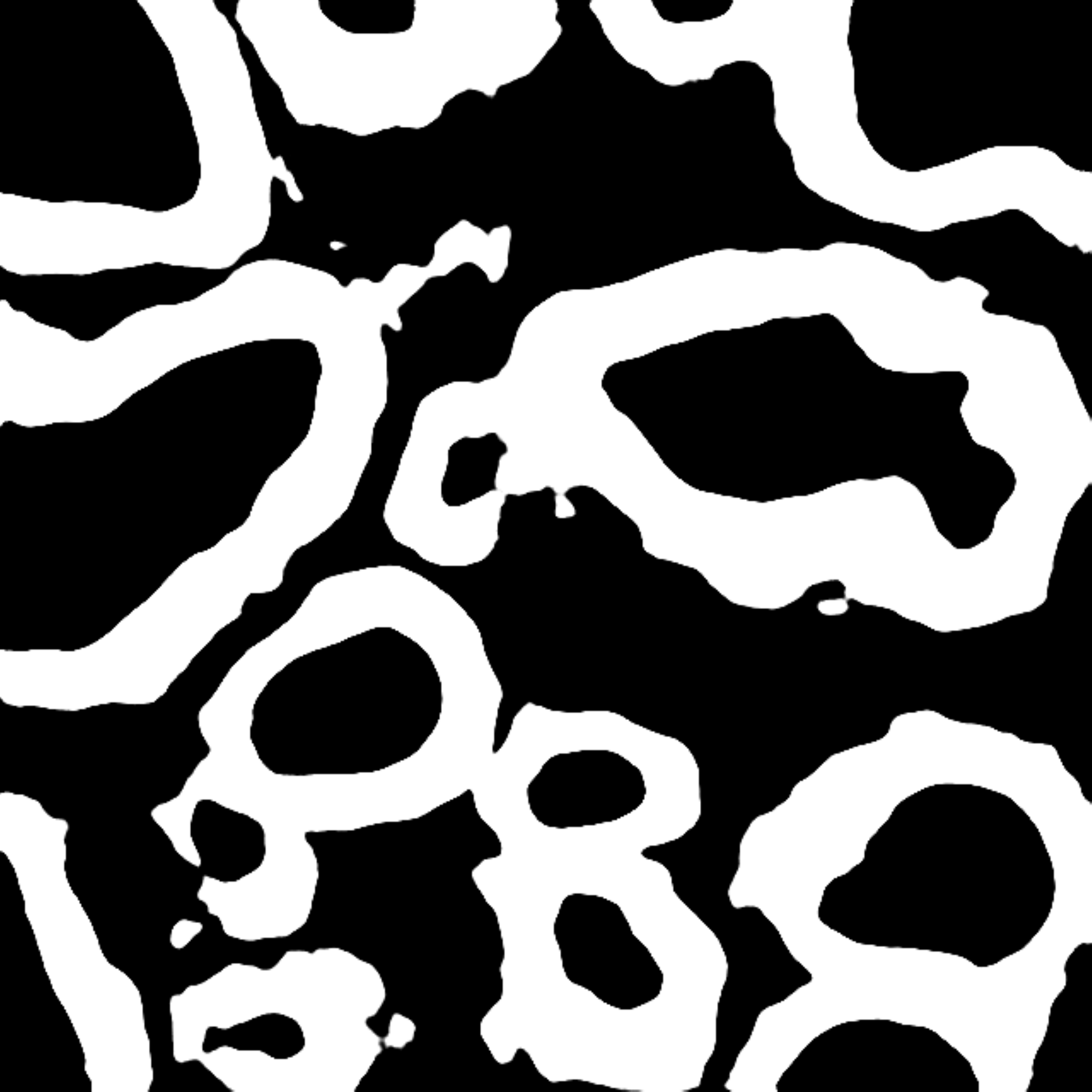}
\tcbincludegraphics{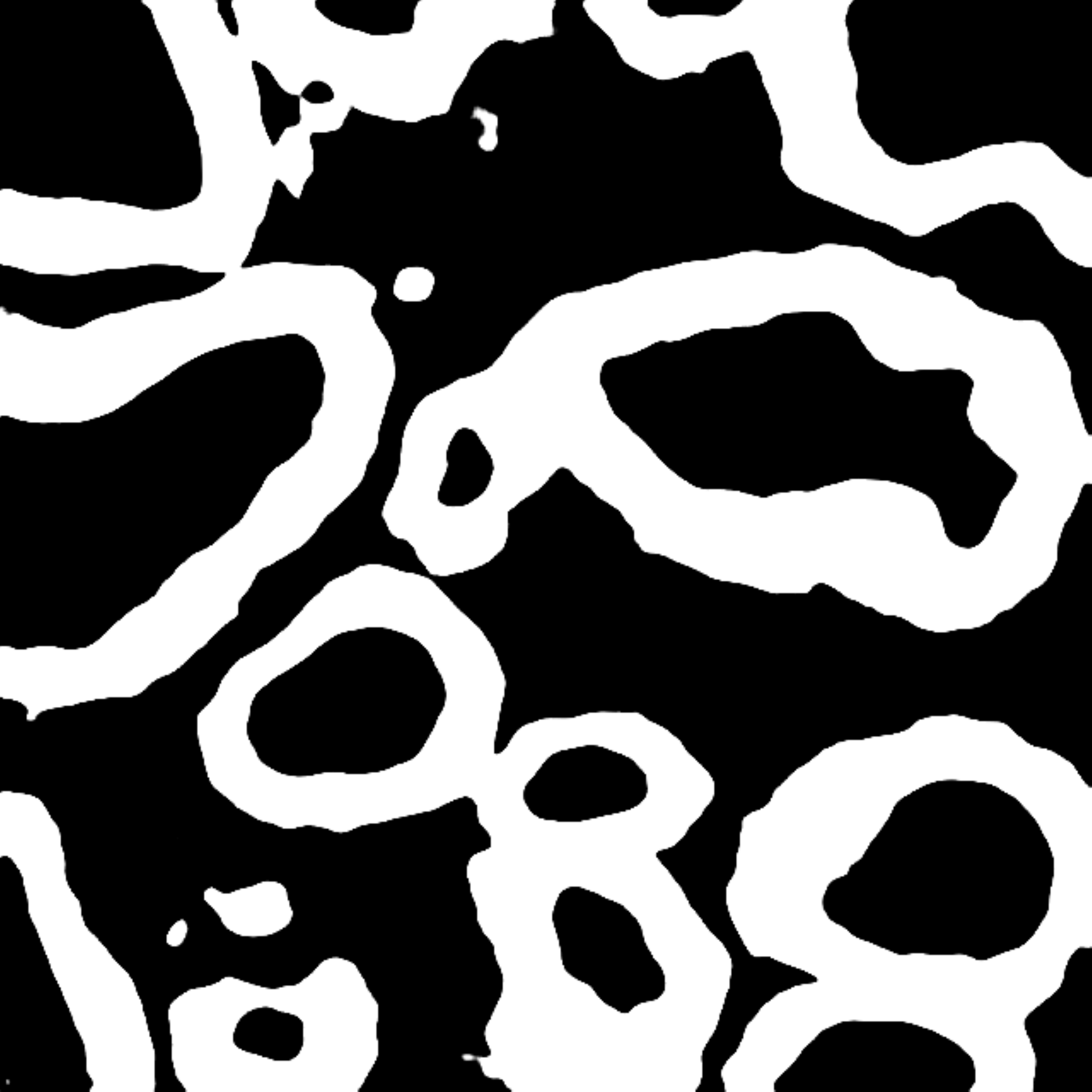}
\tcbincludegraphics{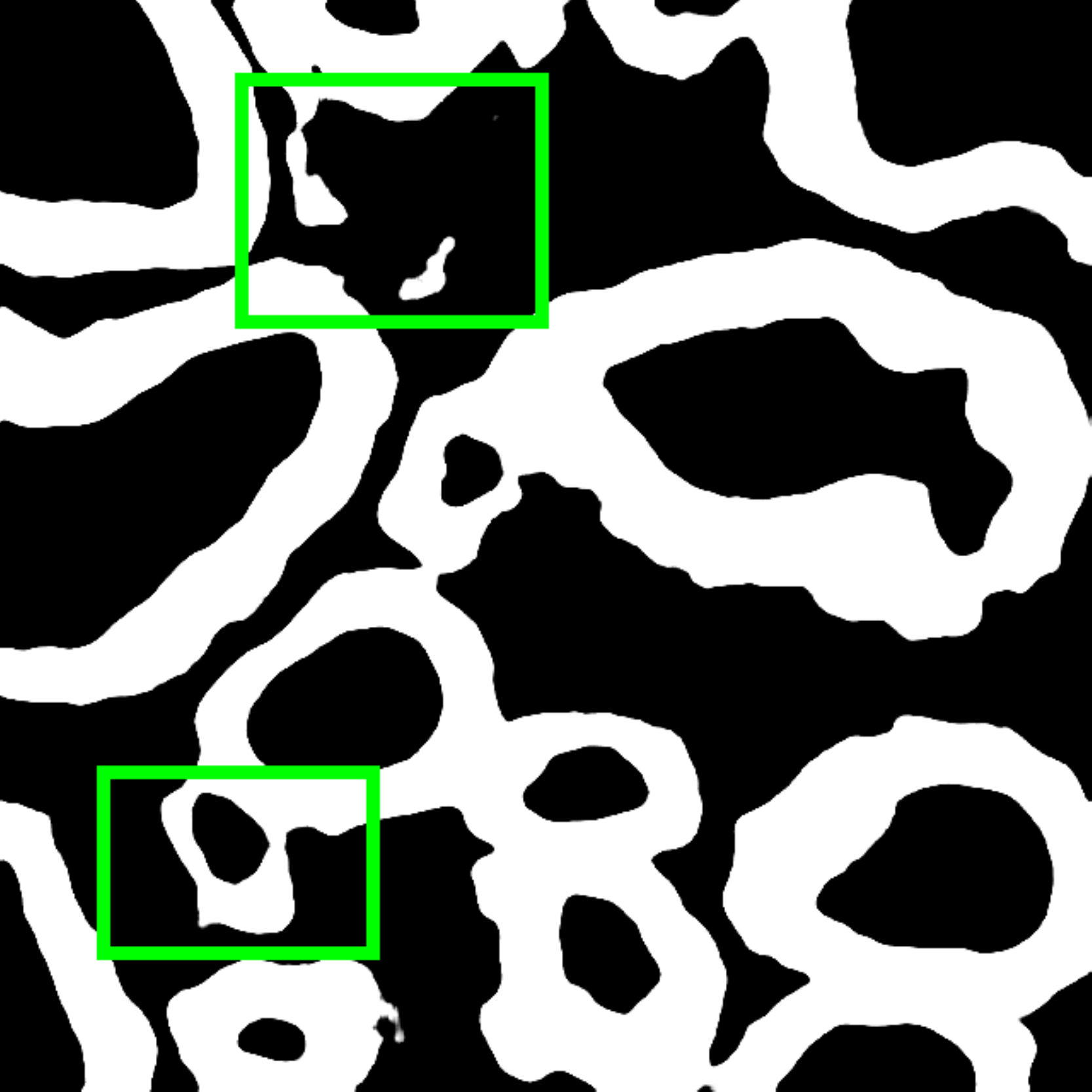}
\end{tcbraster}
\begin{tcbraster}[raster columns=3, raster equal height,
raster column skip=10pt, raster row skip=4pt, raster every box/.style={blank}]
\tcbincludegraphics{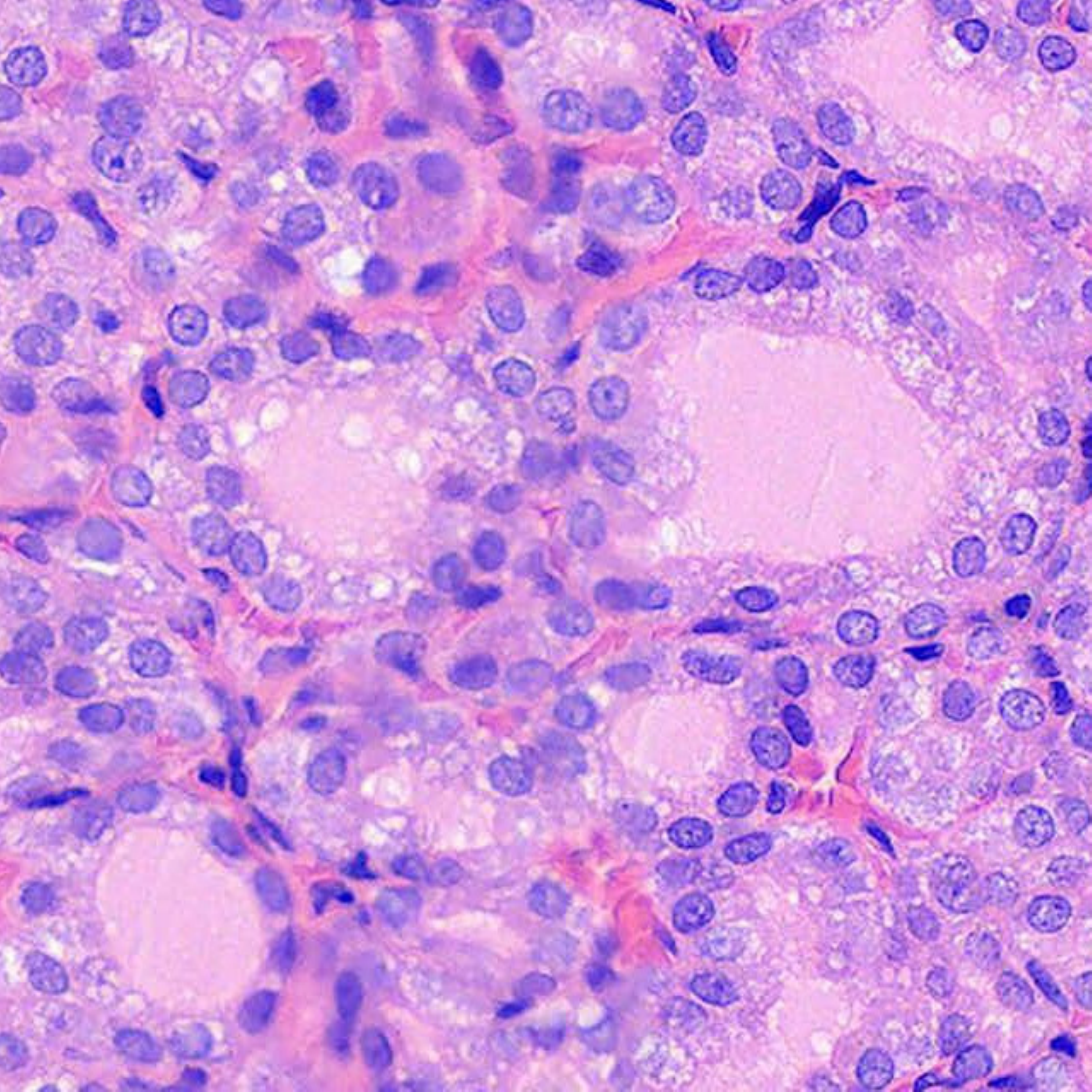}
\tcbincludegraphics{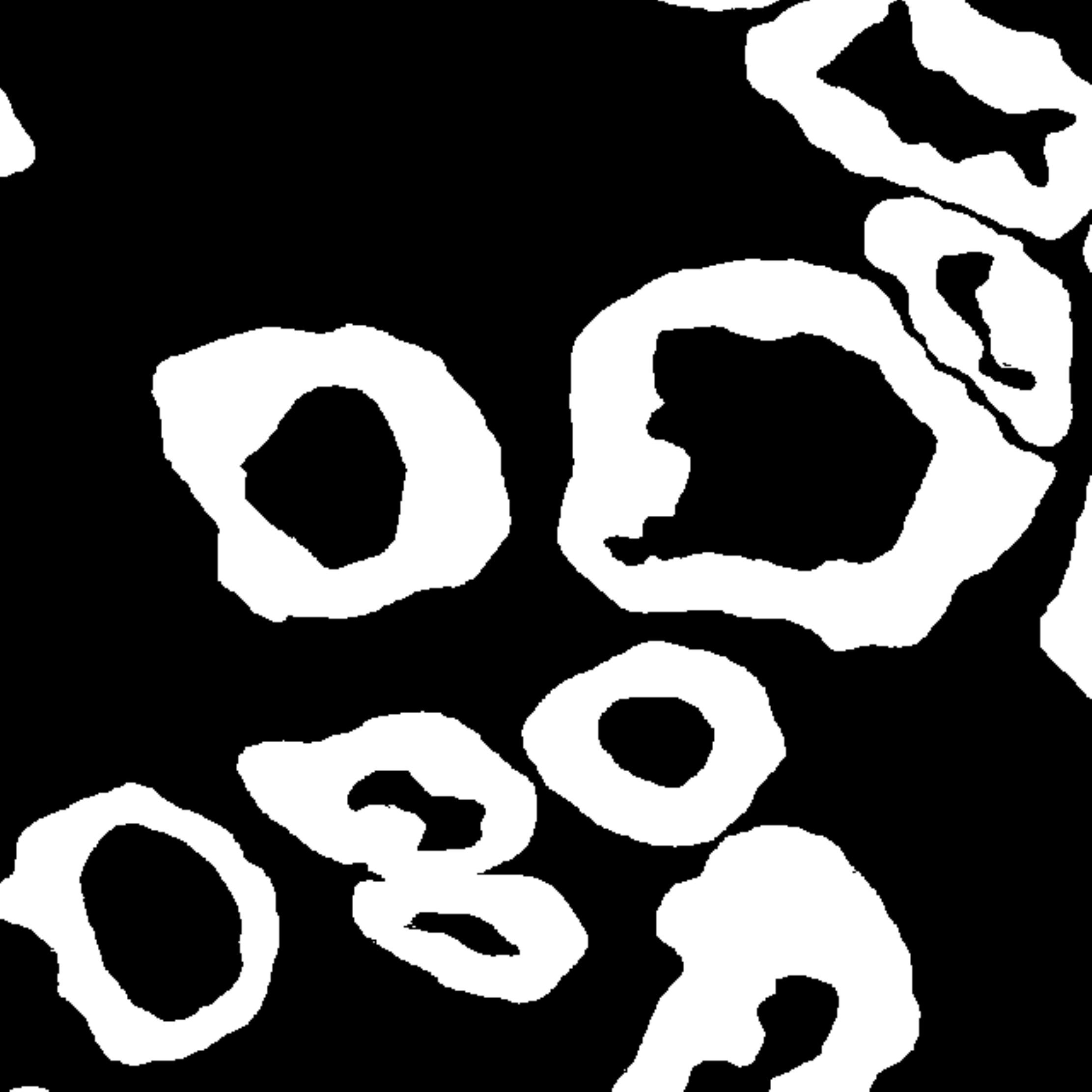}
\tcbincludegraphics{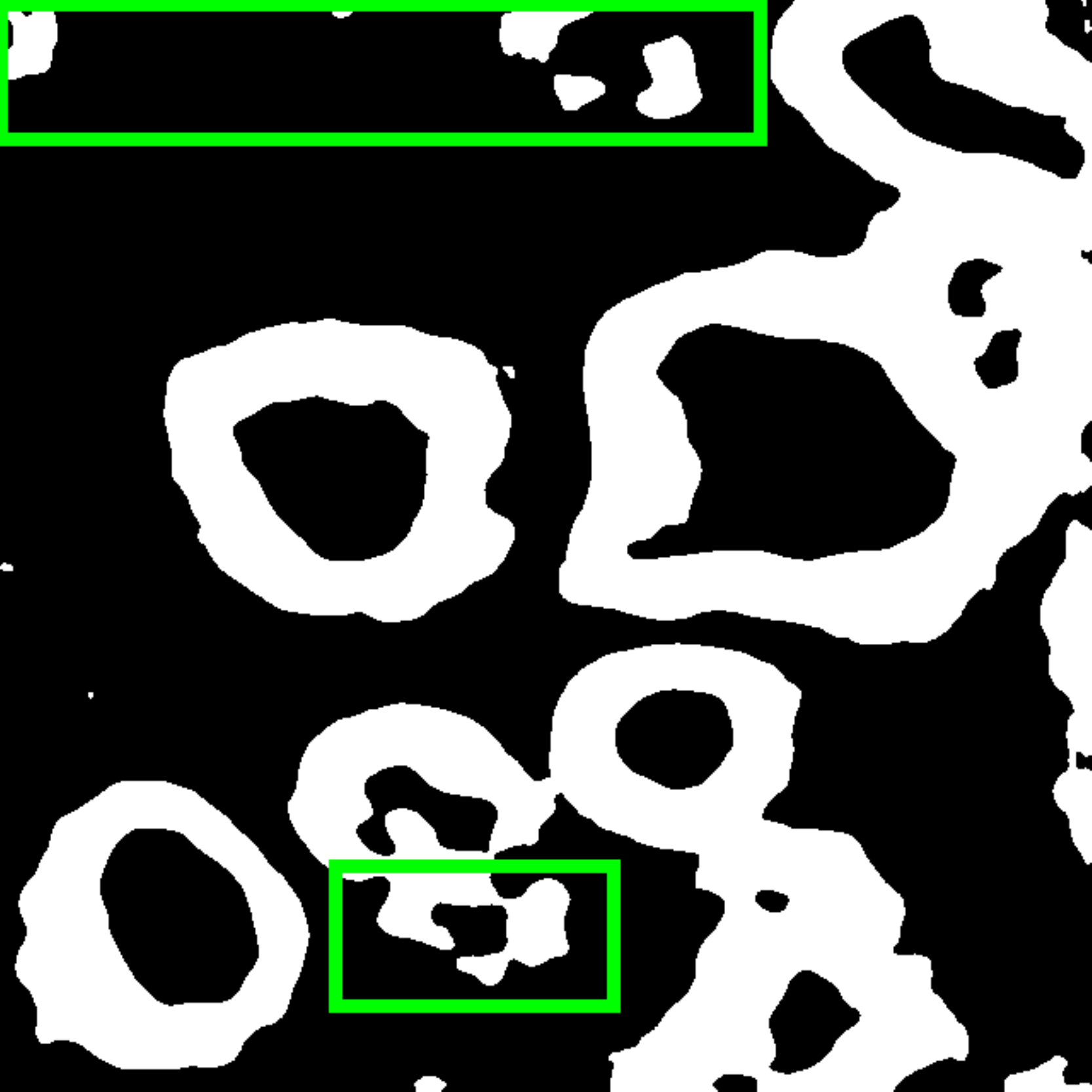}
\tcbincludegraphics{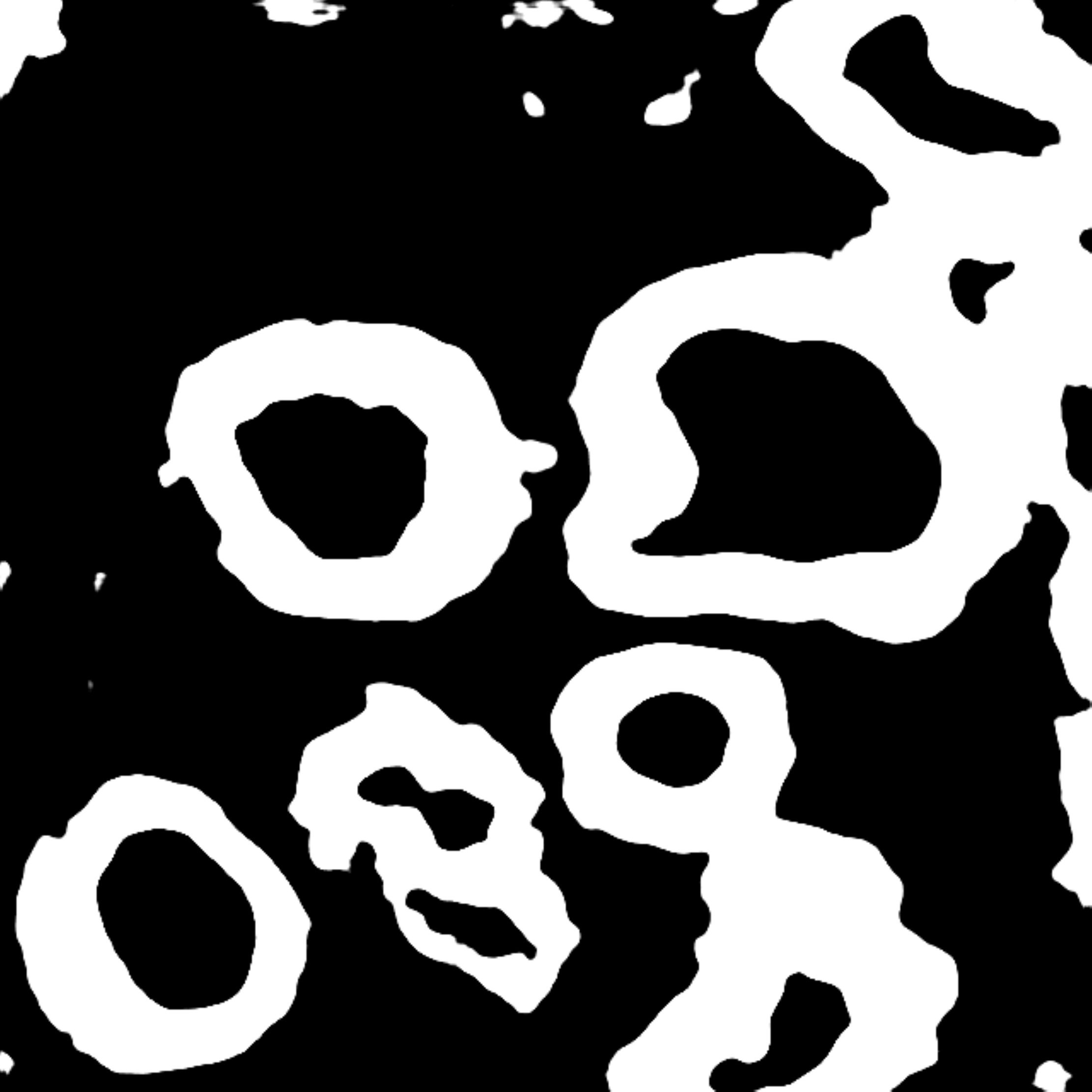}
\tcbincludegraphics{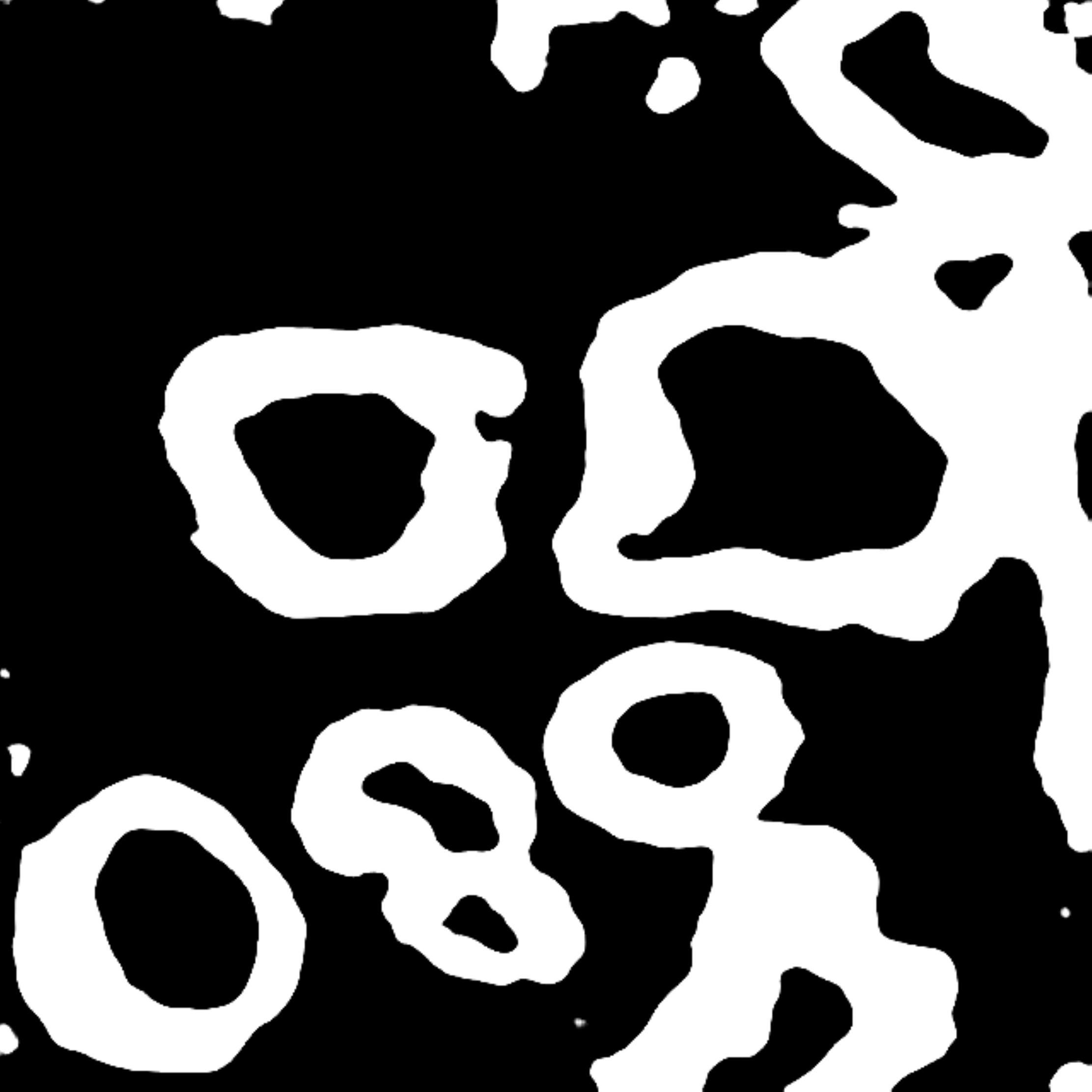}
\tcbincludegraphics{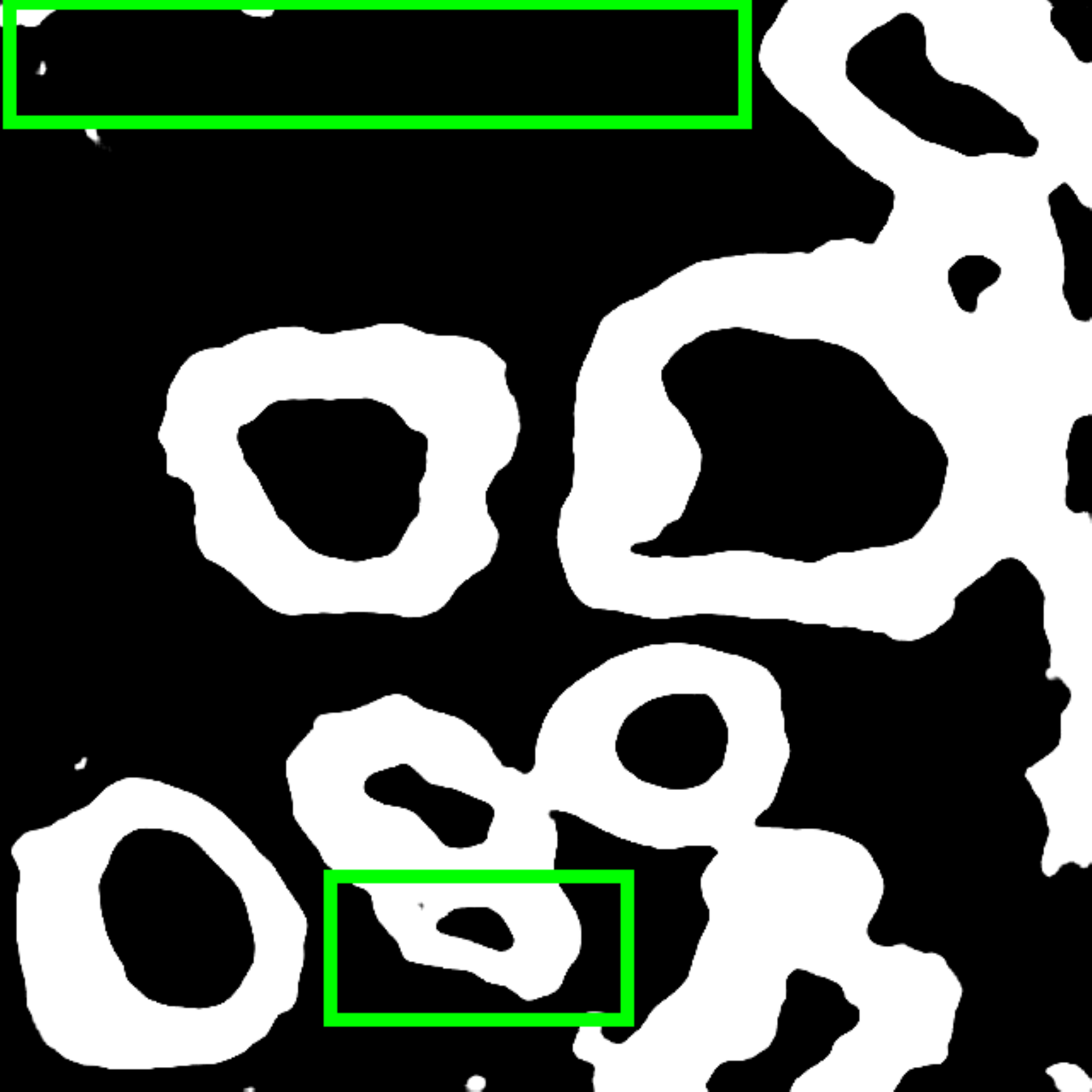}
\end{tcbraster}
\caption{Two image examples and their segmentations. First row:
  input image, ground-truth, and result from U-Net. Second row: results
  from Res-Seg-Net$_\textrm{horz}$, Res-Seg-Net$_\textrm{non-fixed}$,
  and Res-Seg-Net$_\textrm{fixed}$. Rows 3 and 4 show another example.}
\label{fig:results_3}
\end{figure}

\begin{figure}[htb]
\centering
\begin{tcbraster}[raster columns=3, raster equal height,
raster column skip=10pt, raster row skip=4pt, raster every box/.style={blank}]
\tcbincludegraphics{origin_2.pdf}
\tcbincludegraphics{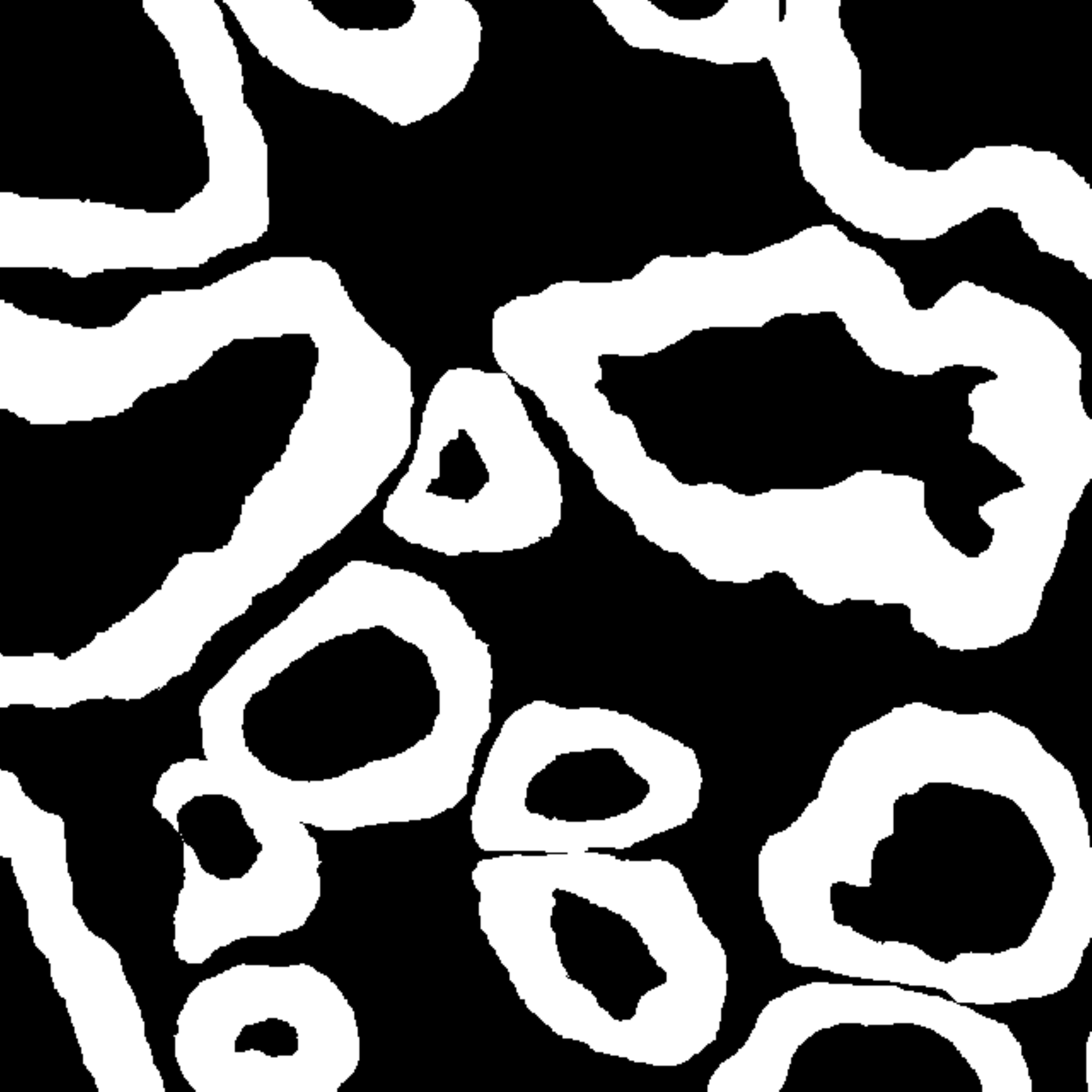}
\tcbincludegraphics{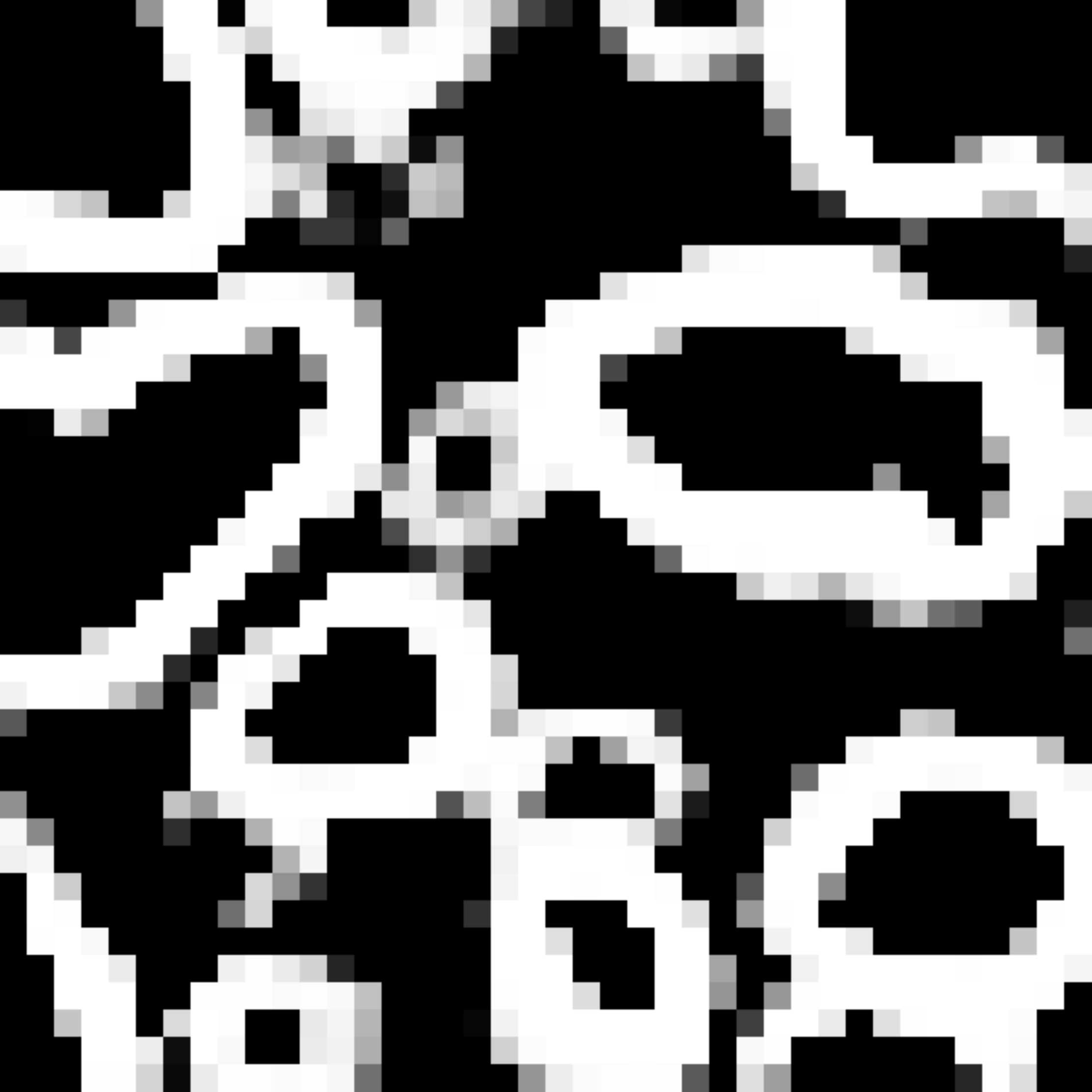}
\tcbincludegraphics{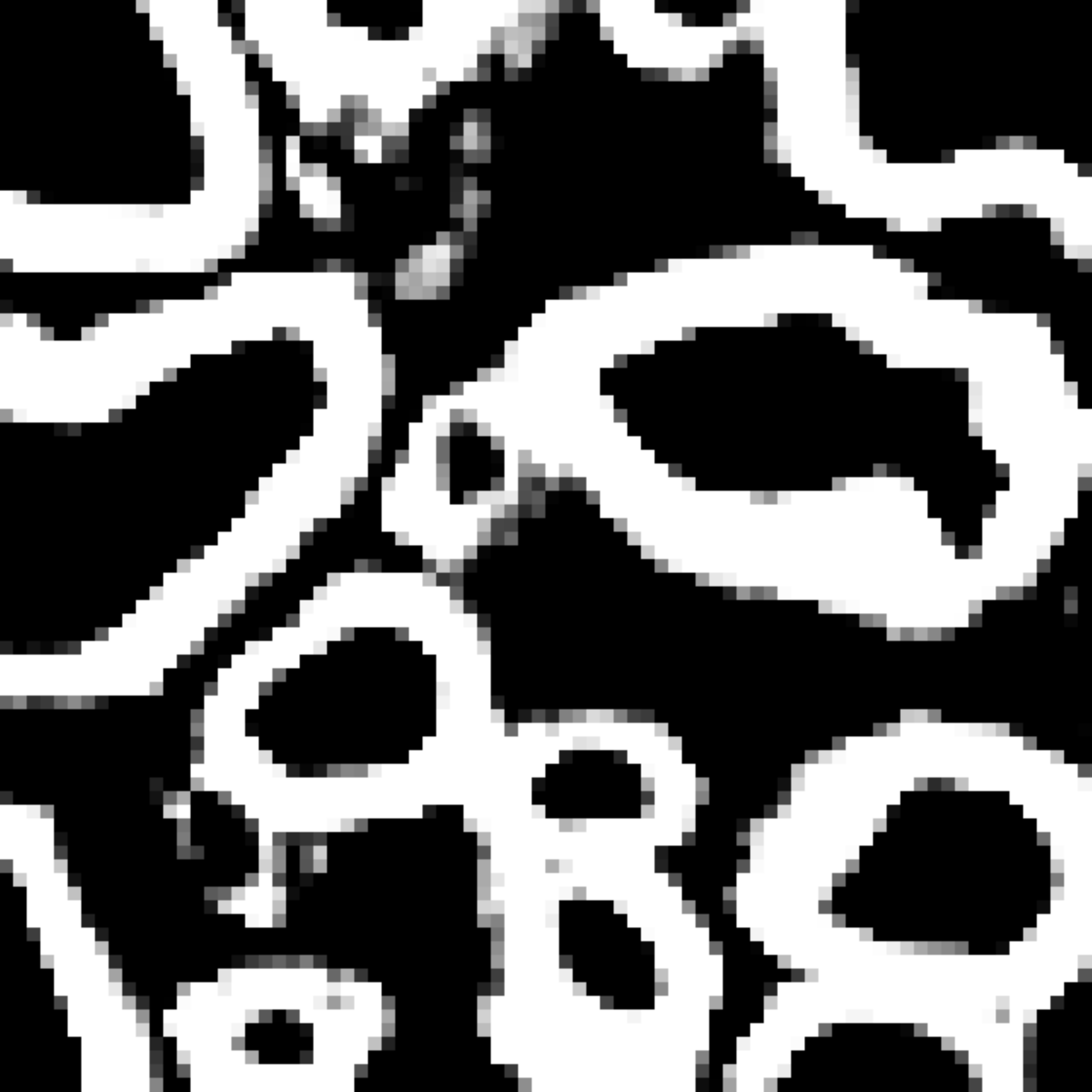}
\tcbincludegraphics{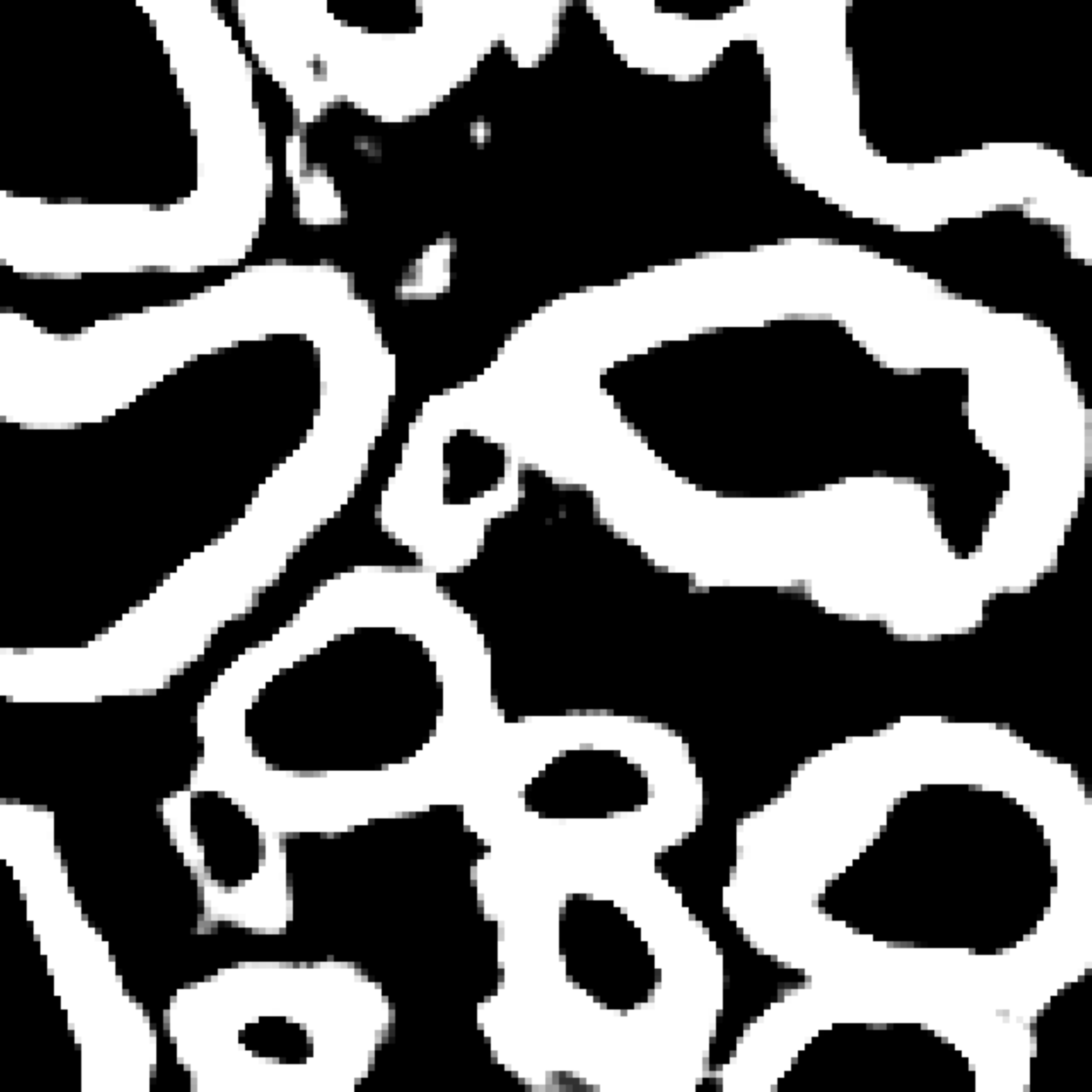}
\tcbincludegraphics{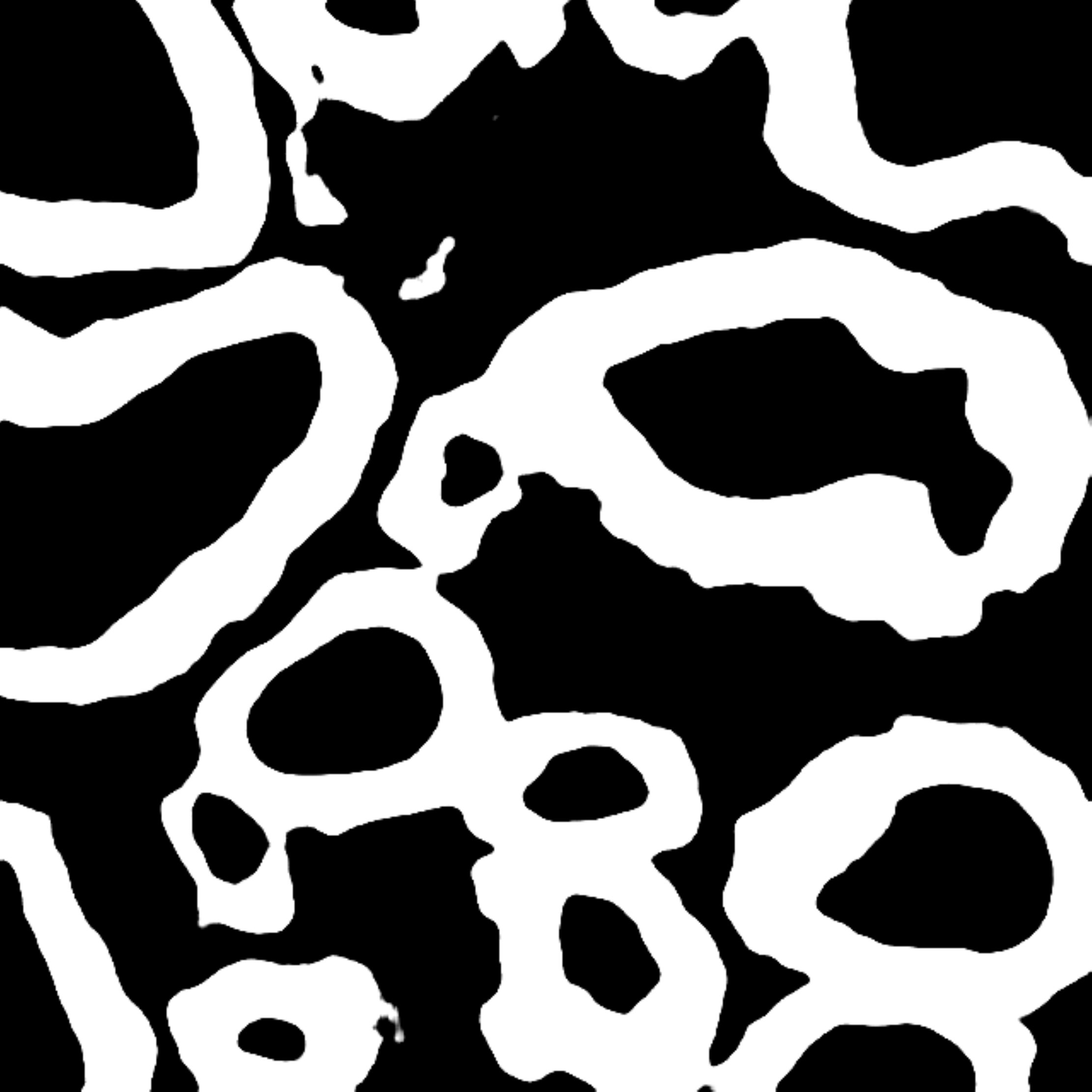}
\end{tcbraster}
\caption{Segmentation results of Res-Seg-Net$_\textrm{fixed}$. Top
  row: input, ground-truth, first level segmentation. Bottom row:
  results from the second, third and fifth levels. }
\label{fig:results_4}
\end{figure}


{\bf{Results}} As we mentioned in section 2.1, we intend to explore two different weight updating schemes in our Res-Seg-Net. One version is to fix upsampled prob-map after it is sent to the upper layer. We call this version Res-Seg-Net$_\textrm{fixed}$. The other setup allows all weights to be updated freely, which we term Res-Seg-Net$_\textrm{non-fixed}$.

To explore the effectiveness of the notion of vertical refinements, i.e., residual updates along coarse-fine hierarchy, we also designed a solution of horizontal refinements as a competing model. More specifically, we stack five Res-Seg modules with upsampling removed, only at the last layer of the modified U-Net, which would carry out segmentation refining only along the highest resolution. We name this model Res-Seg-Net$_\textrm{horz}$. The weights updating scheme in Res-Seg-Net$_\textrm{horz}$ is the fixed version -- prob-maps sent to the upper layers are all set to untrainable.

\begin{table}
  \caption{Segmentation results on Follicles}
\vspace{0.05in}
\centering
\scalebox{0.9}{
\begin{tabular}{c|cc}
  \hline
  \hline
\multicolumn{1}{c|}{\textbf{FCN}} &
\multicolumn{2}{c}{\textbf{Dataset}}\\
\cline{2-3}
\multicolumn{1}{c}{\textbf{ }}&
\multicolumn{1}{|c}{\textbf{Validation}} &
\multicolumn{1}{c}{\textbf{Test}}\\
\hline

  \multicolumn{1}{c|}{\text{Res-Seg-Net$_\textrm{fixed}$}}& 86.44&{\bf{85.51}} \\
  \cline{1-3}
  \multicolumn{1}{c|}{\text{Res-Seg-Net$_\textrm{non-fixed}$}}&86.67&84.97 \\
  \cline{1-3}
  \multicolumn{1}{c|}{\text{Res-Seg-Net$_\textrm{horz}$}}&86.19&85.23\\
  \cline{1-3}
  \multicolumn{1}{c|}{\text{UNet (Baseline)}}& 86.11&84.79 \\
  \hline
\end{tabular}
}
\label{T:follicle}
\end{table}

The results of the four competing models are shown in
Table~\ref{T:follicle}. The {\it{Validation}} column shows the best
result of each model on the validation set, while the {\it{Test}}
column contains the results on the test dataset. The results show that
all the Res-Seg based models outperform the U-Net, where
Res-Seg-Net$_\textrm{fixed}$ obtained the highest DSC on the test
data.

Fig.~\ref{fig:results_3} shows two image examples, their ground-truth
masks, and the final segmentations generated by competing
models. Comparing with U-Net, Res-Seg based models generally have
fewer false positives. Comparing with Res-Seg-Net$_\textrm{horz}$,
Res-Seg-Net$_\textrm{fixed}$ and Res-Seg-Net$_\textrm{non-fixed}$ both
generate cleaner outputs, which can serve as an evidence that
hierarchical refinings are effective in improving segmentations in
both accuracy and robustness. These effects can be clearly seen within
the areas highlighted with green boxes. 

We fed Res-Seg-Net$_\textrm{fixed}$ with the patch in the first row of
Fig.~\ref{fig:results_3}, and output their segmentations maps of each
layer in Fig.~\ref{fig:results_4}.  The segmentation refining process
is evident, as more and more details are added to the finer
outputs. The low resolution segmentations tend to catch the primary
shapes of the target objects. Moving upwards, they not only provide
guidance for fine-resolution labelings to capture more details, but
also set up certain guard to reduce the appearance of noisy spots.

\vspace{-0.1in}
\section{Conclusions}

The Res-Seg module proposed in this paper facilitates the
multi-resolution processing of hierarchical presentations and
information flow in a network. Mounted onto the expansion path of an
FCN, it can help each layer focus on learning incremental refinements
from its previous layer. Both primary object shapes and boundary
details can be potentially better captured through the valuable
mechanism brought by Res-Seg-Net. Exploring applications of
Res-Seg-Net on more datasets, as well as the potential integrations of
Res-Seg module with deep neural networks in other application areas,
e.g., detection, are the directions of our future efforts.

\newpage

\small

\bibliographystyle{IEEEbib}
\bibliography{zhewei_isbi_2019}

\end{document}